\documentstyle{article}

\title{Assessing agreement on classification tasks: 
the kappa statistic}

\author{Jean Carletta%
  \thanks{Human Communication Research Centre, 
        University of Edinburgh, 2 Buccleuch Place,
    Edinburgh EH8 9LW, Scotland}}

\begin{document}

\maketitle
                                                           
\begin{abstract} 
Currently, computational linguists and cognitive scientists 
working in the area
of discourse and dialogue argue that their subjective
judgments are reliable using several different statistics, none of
which are easily interpretable or comparable to each other.
Meanwhile, researchers in content analysis have already experienced
the same difficulties and come up with a solution in the kappa
statistic.  We discuss what is wrong with reliability measures as
they are currently used for discourse and dialogue work in
computational linguistics and cognitive science, and argue that
we would be better off as a field adopting techniques from content
analysis.
\end{abstract}

\section{Introduction}

Computational linguistic and cognitive science work on discourse
and dialogue relies on subjective judgments.  For instance, much
current research on discourse phenomena distinguishes between
behaviours which tend to occur at or around discourse segment
boundaries and those which do not \cite{ACL93:Passonneau&Litman},
\cite{HCRC:Kowtko-etal}, \cite{COLING90:Litman&Hirschberg}
\cite{prosody:Cahn}.  Although in some cases discourse segments
are defined automatically (e.g., Rodrigues and Lopes'
\cite{COLING92:Rodrigues&Lopes} definition based on temporal
relationships), more usually discourse segments are defined
subjectively based on the intentional structure of the discourse,
and then other phenomena are related to them.  At one time, it
was considered sufficient when working with such judgments to
show examples based on the authors' interpretation
(paradigmatically, \cite{Grosz&Sidner-CL}, but also countless
others).  Research was judged according to whether or not the
reader found the explanation plausible.  Now, researchers are
beginning to require evidence that people besides the authors
themselves can understand and make the judgments underlying the
research reliably.  This is a reasonable requirement because if
researchers can't even show that different people can agree about
the judgments on which their research is based, then there is no
chance of replicating the research results.  Unfortunately, as a
field we have not yet come to agreement about how to show
reliability of judgments.  For instance, consider the following
arguments for reliability.  We have chosen these examples both
for the clarity of their arguments and because taken as a set
they introduce the full range of issues we wish to discuss.

\begin{description}

\item[(1)] Kowtko et al. \cite{HCRC:Kowtko-etal}, in arguing that it is
  possible to mark conversational move boundaries, cite separately for
  each of three naive coders the ratio of the number of times they
  agreed with an ``expert'' coder about the existence of a boundary
  over the number of times either the naive coder or the expert marked
  a boundary.  She does not describe any restrictions on possible
  boundary sites.

\item[(2)] Once conversational move boundaries have been marked on a
  transcript, Kowtko et al. argue that naive coders can reliably
  place moves into one of thirteen exclusive categories by citing
  pairwise agreement percentages figured over all thirteen categories,
  again looking at each of the three naive coders separately.  Litman
  and Hirschberg \cite{COLING90:Litman&Hirschberg} use this same
  pairwise technique for assessing the reliability of cue phrase
  categorisation, using two equal-status coders and three categories.

\item[(3)] Silverman et al. \cite{ICSLP92:TOBI}, in arguing that sets
  of coders can agree on a range of category distinctions involved in
  the TOBI system for labelling English prosody, cite the ratio of
  observed agreements over possible agreements, measuring over all
  possible pairings of the coders.  For example, they use this measure
  for determining the reliability of the existence and category of
  pitch accents, phrase accents, and boundary tones.  They measure
  agreement over both a pool of highly experienced coders and a larger
  pool of mixed-experience coders, and argue informally that since the
  level of agreement is not much different between the two, their
  coding system is easy to learn.

\item[(4)] Passonneau and Litman \cite{ACL93:Passonneau&Litman}, in
  arguing that naive subjects can reliably agree on whether or not
  given prosodic phrase boundaries are also discourse segment
  boundaries, measure reliability using `percent agreement', defined
  as the ratio of observed agreements with the majority opinion among
  seven naive coders to possible agreements with the majority opinion.

\end{description}

Although (1) and Kowtko's use of (2) differ slightly from Litman and
Hirschberg's use of (2), (3) and (4) in clearly designating one coder
as an ``expert'', all these studies have $n$ coders place some kind of
units into $m$ exclusive categories. Note that the cases of testing
for the existence of a boundary can be treated as coding ``yes'' and
``no'' categories for each of the possible boundary sites; this
treatment is used by measures (3) and (4) but not by measure (1).  All
four approaches seem reasonable when taken at face value. However, the
four measures of reliability bear no relationship to each other.
Worse yet, since none of them take into account the level of agreement
one would expect coders to reach by chance, none of them are
interpretable even on their own.  We first explain what effect chance
expected agreement has on each of these measures, and then argue that
we should adopt the kappa statistic \cite{Siegel&Castellan} as a
uniform measure of reliability.

\section{Chance expected agreement}

Measure (2) seems a natural choice when there are two coders, and
there are several possible extensions when there are more coders,
including citing separate agreement figures for each important pairing
(as Kowtko does by designating an expert), counting a unit as agreed
only if all coders agree on it, or measuring one agreement over all
possible pairs of coders thrown in together.  Taking just the two
coder case, the amount of agreement we would expect coders to reach by
chance depends on the number and relative proportions of the
categories used by the coders.  For instance, consider what happens
when the coders randomly place units into categories instead of using
an established coding scheme.  If there are two categories which occur
in equal proportions, on average they would agree with each other half
of the time; each time the second coder makes a choice, there is a
fifty/fifty chance of coming up with the same category as the first
coder.  If instead, the two coders were to use four categories in
equal proportions, we would expect them to agree 25\% of the time
(since no matter what the first coder chooses, there is a 25\% chance
that the second coder will agree.)  And if both coders were to use one
of two categories, but use one of the categories 95\% of the time, we
would expect them to agree 90.5\% of the time ($.95^{2} + .05^{2}$,
or, in words, 95\% of the time the first coder chooses the first
category, with a .95 chance of the second coder also choosing that
category, and 5\% of the time the first coder chooses the second
category, with a .05 chance of the second coder also doing so).  This
makes it impossible to interpret raw agreement figures using measure
(2).  This same problem affects all of the possible ways of extending
measure (2) to more than two coders.

Now consider measure (3), which has an advantage over measure (2) when
there is a pool of coders, none of whom should be distinguished, in
that it produces one figure which sums reliability over all coder
pairs.  Measure (3) still falls foul of the same problem with expected
chance agreement as measure (2) because it does not take into account
the number of categories which occur in the coding scheme.

Measure (4) is a different approach to measuring over multiple
undifferentiated coders.  Note that although Passonneau and Litman are
looking at the presence or absence of discourse segment boundaries,
measure (4) takes into account agreement that a prosodic phrase
boundary is not a discourse segment boundary, and therefore treats the
problem as a two-category distinction.  Measure (4) falls foul of the
same basic problem with chance agreement as measures (2) and (3), but
in addition, the statistic itself guarantees at least 50\% agreement
by only pairing off coders against the majority opinion.  It also
introduces an ``expert'' coder by the back door in assuming that the
majority is always right, although this stance is somewhat at odds
with Passonneau and Litman's subsequent assessment of a boundary's
strength from one to seven based on the number of coders who noticed
it.

Measure (1) looks at almost exactly the same type of problem as
measure (4), the presence or absence of some kind of boundary.
However, since one coder is explicitly designated as an ``expert'', it
doesn't treat the problem as a two category distinction, but looks
only at cases where either coder marked a boundary as present.
Without knowing the density of conversational move boundaries in the
corpus, this makes it difficult to assess how well the coders agreed
on the absence of boundaries and to compare measures (1) and (4).  In
addition, note that since false positives and missed negatives are
rolled together in the denominator of the figure, measure (1) does not
really distinguish expert and naive coder roles as much as it might.
However this style of measure does have some advantages over measures
(2), (3), and (4), since these measures produce artificially high
agreement figures when one category of a set predominates, as is the
case with boundary judgments.  One would expect measure (1)'s results
to be high under any circumstances, and it is not affected by the
density of boundaries.

So far, we have shown that all four of these measures produce figures
which are at best, uninterpretable and at worst, misleading.  Kowtko
et al. make no comment about the meaning of their figures other than
to say that the amount of agreement they show is reasonable; Silverman
et al. simply point out that where figures are calculated over
different numbers of categories, they are not comparable.  On the
other hand, Passonneau and Litman note that their figures are not
properly interpretable and attempt to overcome this failing to some
extent by showing that the agreement which they have obtained at least
significantly differs from random agreement.  Their method for showing
this is complex and of no concern to us here, since all it tells us is
that it is safe to assume that the coders were not coding randomly ---
reassuring, but no guarantee of reliability.  It is more important
to ask {\em how different} the results are from random and whether or
not the data produced by coding is too noisy to use for the purpose
for which it was collected.

\section{The kappa statistic}

The concerns of these researchers are largely the same as those in the
field of content analysis (see especially \cite{Krippendorff} and
\cite{Weber}), which has been through the same problems as we are
currently facing and in which strong arguments have been made for
using the kappa coefficient of agreement \cite{Siegel&Castellan} as a
measure of reliability.\footnote{There are several variants of the
  kappa coefficient in the literature, including one, Scott's {\it
    pi}, which actually has been used at least once in our field, to
  assess agreement on move boundaries in monologues using action
  assembly theory \cite{Greene&Cappella}.  Krippendorff's $\alpha$ is
  more general than Siegel and Castellan's $K$ in that Krippendorff
  extends the argument from category data to interval and ratio
  scales; this extension might be useful for, for instance, judging
  the reliability of TOBI break index coding, since some researchers
  treat these codes as inherently scalar \cite{ICSLP92:TOBI}.
  Krippendorff's $\alpha$ and Siegel and Castellan's $K$ actually
  differ slightly when used on category judgments in the assumptions
  under which expected agreement is calculated.  Here we use Siegel
  and Castellan's $K$ because they explain their statistic more
  clearly, but the value of $\alpha$ is so closely related, especially
  under the usual expectations for reliability studies, that
  Krippendorff's statements about $\alpha$ hold, and we conflate the
  two under the more general name `kappa'.  The advantages and
  disadvantages of different forms and extensions of kappa have been
  discussed in many fields but especially in medicine; see, for
  example, \cite{JAMA:Berry,JAMA:Goldman,Kraemer,Soeken&Prescott}.}

The kappa coefficient ($K$) measures pairwise agreement
among a set of coders making category judgments, correcting for
expected chance agreement.
\[
K = \frac{P(A) - P(E)}{1-P(E)}
\] where $P(A)$ is the proportion of times that the coders agree and
$P(E)$ is the proportion of times that we would expect them to agree
by chance, calculated along the lines of the intuitive argument
presented above.  (For complete instructions on how to calculate $K$,
see \cite{Siegel&Castellan}.)  When there is no agreement other than
that which would be expected by chance $K$ is zero. When there is
total agreement, $K$ is one. When it is useful to do so, it is
possible to test whether or not $K$ is significantly different from
chance, but more importantly, interpretation of the scale of agreement
is possible.  Krippendorff \cite{Krippendorff} discusses what makes an
acceptable level of agreement, while giving the caveat that it depends
entirely on what one intends to do with the coding.  For instance, he
claims that it is often impossible to find associations between two
variables which both rely on coding schemes with $K<.7$, and says that
content analysis researchers generally think of $K>.8$ as good
reliability, with $.67<K<.8$ allowing tentative conclusions to be
drawn.  We would add two further caveats. First, although kappa
addresses many of the problems we have been struggling with as a field,
in order to compare $K$ across studies, the underlying assumptions
governing the calculation of chance expected agreement still require
the units over which coding is performed to be chosen sensibly and
comparably.  (To see this, compare, for instance, what would happen to
the statistic if the same discourse boundary agreement data were
calculated variously over a base of clause boundaries, transcribed
word boundaries, and transcribed phoneme boundaries.)  Where no
sensible choice of unit is available pretheoretically, measure (1) may
still be preferred. Secondly, coding discourse and dialogue phenomena,
and especially coding segment boundaries, may be inherently more
difficult than many previous types of content analysis (for instance,
dividing newspaper articles based on subject matter).  Whether we have
reached (or will be able to reach) reasonable level of agreements in
our work as a field remains to be seen; our point here is merely that
if as a community we adopt clearer statistics, we will be able to
compare results in a standard way across different coding schemes and
experiments and to evaluate current developments, and that will
illuminate both our individual results and the way forward.

\section{Expert versus naive coders}

In assessing the amount of agreement among coders of category
distinctions, the kappa statistic normalises for the amount of
expected chance agreement and allows a single measure to be calculated
over multiple coders.  This makes it applicable to the studies which
we have described, plus more besides.  However, we have yet to discuss
the role of expert coders in such studies.  Kowtko designates one
particular coder as the expert.  Passonneau and Litman have only naive
coders, but in essence have an expert opinion available on each unit
classified in terms of the majority opinion.  Silverman et al. treat
all coders indistinguishably, although they do build an interesting
argument about how agreement levels shift when a number of less
experienced transcribers are added to a pool of highly experienced
ones.  We would argue that in subjective codings such as these, there
are no real experts.  We concur with Krippendorff that what counts is
how totally naive coders manage based on written instructions.
Comparing naive and expert coding as Kowtko does can be a useful
exercise, but instead of assessing the naive coders' accuracy, it
really measures how well the instructions convey what she thinks they
do.  (Krippendorff gives well-established techniques which generalise
on this sort of ``odd-man-out'' result, which involve isolating
particular coders, categories, and kinds of units in order to
establish where any disagreement is coming from.)  In Passonneau and
Litman, the reason for comparing to the majority opinion is less
clear.

Despite our argument, there {\em are} occasions when one opinion
should be treated as the expert one.  For instance, one can imagine
determining whether coders using a simplified coding scheme match what
can be obtained by some better but more expensive method, which might
itself be either objective or subjective.  In these cases, we would
argue that it is still appropriate to use a variation on the kappa
statistic which only looks at pairings of agreement with the expert
opinion rather than looking at all possible pairs of coders.  This
could be done by interpreting $P(A)$ as the proportion of times that
the ``naive'' coders agree with the expert one and $P(E)$ as the
proportion of times we would expect the naive coders to agree with the
expert one by chance.

\section{Conclusions}

We have shown that existing measures of reliability in discourse and
dialogue work are difficult to interpret, and we have suggested a
replacement measure, the kappa statistic, which has a number of
advantages over these measures.  Kappa is widely accepted in the field
of content analysis.  It is interpretable, allows different results to
be compared, and suggests a set of diagnostics in cases where the
reliability results are not good enough for the required purpose.  We
suggest that this measure be adopted more widely within our own
research community.

\section{Author Note}

This work was supported by grant number G9111013 of the U.K. Joint
Councils Initiative in Cognitive Science and Human-Computer
Interaction and an Interdisciplinary Research Centre Grant from the
Economic and Social Research Council (U.K.) to the Universities of
Edinburgh and Glasgow.

\pagebreak
\bibliographystyle{plain}

\newcommand{\etalchar}[1]{$^{#1}$}

\end{document}